\documentclass[journal]{IEEEtran}

\ifCLASSINFOpdf
\else
\fi

\usepackage{graphicx}
\usepackage{amssymb}

\usepackage[utf8]{inputenc} 
\usepackage[T1]{fontenc}    
\usepackage{hyperref}       
\usepackage{url}            
\usepackage{booktabs}       
\usepackage{amsfonts}       
\usepackage{nicefrac}       
\usepackage{microtype}      
\usepackage{graphicx}
\usepackage{amsthm}
\usepackage{amsmath}
\usepackage[noend]{algpseudocode}
\usepackage[english]{babel}
\usepackage[T1]{fontenc}
\usepackage[utf8]{inputenc}
\usepackage{authblk}
\usepackage{bbm}
\usepackage{wrapfig}

\usepackage{adjustbox}

\usepackage{hyperref}
\usepackage{url}
\usepackage{subcaption}
\usepackage{soul}
\usepackage{ulem}
\usepackage{colortbl}
\usepackage[table]{xcolor}

\newtheorem*{theorem*}{Theorem}
\newtheorem{defn}{Definition}
\newtheorem{prop}{Proposition}
\newtheorem*{prop*}{Proposition}
\newtheorem{conj}{Conjecture}

\usepackage{todonotes}

\begin{document}


\author{Hugo Caselles-Dupr\'e\textsuperscript{1,3}, Michael Garcia-Ortiz\textsuperscript{2}, David Filliat\textsuperscript{1} \\
\textsuperscript{1}Flowers Laboratory (ENSTA Paris \& INRIA), \textsuperscript{2}CitAI, SMCSE, City University of London,\\ \textsuperscript{3}AI Lab (Softbank Robotics Europe)\\
\href{mailto:caselles@ensta.fr}{caselles@ensta.fr}, \href{mailto:mgarciaortiz@softbankrobotics.com}{mgarciaortiz@softbankrobotics.com}, \href{mailto:david.filliat@ensta.fr}{david.filliat@ensta.fr}}

\title{Sensory Commutativity of Action Sequences for Embodied Agents: Theory and Practice}

\maketitle

\begin{abstract}

Perception of artificial agents is one the grand challenges of AI research. Deep Learning and data-driven approaches are successful on constrained problems where perception can be learned using supervision, but do not scale to open-worlds. In such case, for autonomous embodied agents with first-person sensors, 
perception can be learned end-to-end to solve particular tasks. However, literature shows that perception is not a purely passive compression mechanism, and that actions play an important role in the formulation of abstract representations. We propose to study perception for these embodied agents, under the mathematical formalism of group theory in order to make the link between perception and action. In particular, we consider the commutative properties of continuous action sequences with respect to sensory information perceived by such an embodied agent. We introduce the Sensory Commutativity Probability (SCP) criterion which measures how much an agent's degree of freedom affects the environment in embodied scenarios. We show how to compute this criterion in different environments, including realistic robotic setups. We empirically illustrate how SCP and the commutative properties of action sequences can be used to learn about objects in the environment and improve sample-efficiency in Reinforcement Learning.

\end{abstract}


\section{Introduction}

Perception is the medium by which agents organize and interpret sensory stimuli, in order to reason and act in an environment using their available actions \cite{hoffman2018interface}. We focus on scenarios where embodied agents are situated in \textit{realistic} environments, i.e. the agents face partial observability, coherent physics, first-person view with high-dimensional state space, and low-level continuous motor (i.e. action) space with multiple degrees of freedom. 

In classical robotics, we can use a controlled robotic setup where we utilize external information about the agent and the environment, such as position, joint parameters, object positions, and annotated data. This allows the experimenter to distill its knowledge in the form of priors into the system (e.g. knowledge of the workspace in the case of an a robot interacting with objects on a table). However, this information might not be available in the general case. In Nature, children and animals do not have access to this information when they are born. They start from a relatively naive setup, and then build perception via interaction with the environment. We aim at developing theories and applications for this tabula-rasa case where the agent is naive: it can only actuate its motors (without any description of what they do) and receive observations through its sensors. 

Embodied agents, when acting in their environment, produce a stream of sensorimotor data, composed of successions of motor states and sensory information. While most current approaches for building perception consider that the interpretation of sensory information is an isolated problem that only requires extracting relevant information in instantaneous sensor values \cite{mnih2013playing, he2016deep}, several approaches \cite{caselles2019symmetry, laflaquiere2018unsupervised, ghosh2018learning, thomas2017independently} that can be traced back to 1895 \cite{poincare1895espace}, advocate the necessity of studying the relationship between sensors and motors for the emergence of perception.

\begin{figure}[h!]
    \centering
    \includegraphics[scale=0.27]{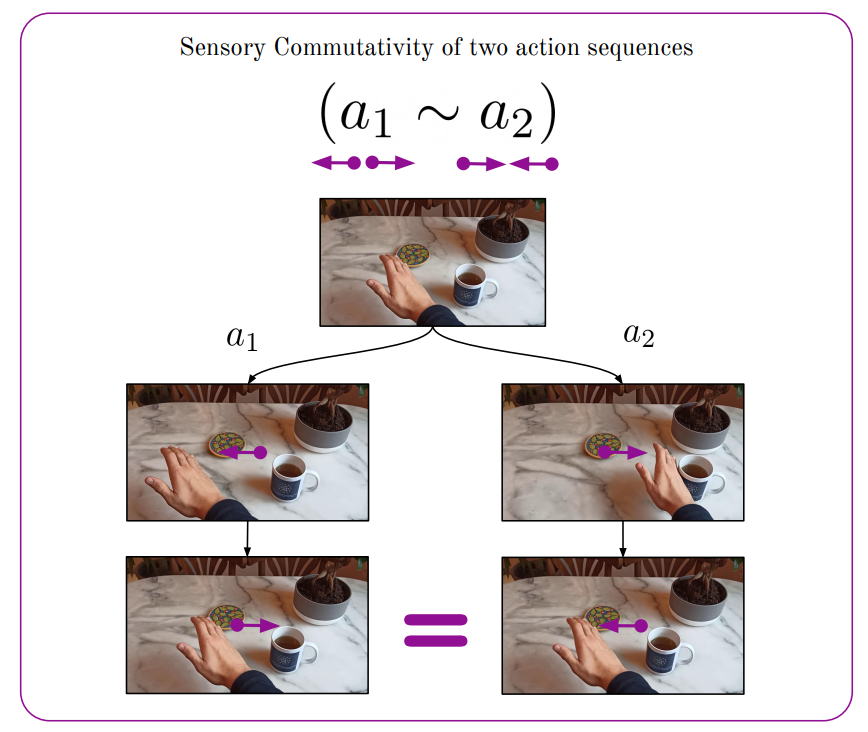}
    \caption{Two action sequences sensory commute if they produce the same sensory state when played in different orders from the same starting position. In this example, the action sequences would not commute if an object would be in the way of the hand movement.}
    \label{fig:scp}
\end{figure}

Inspired by these works, we study the commutativity of action sequences with respect to sensors, which we term sensory commutativity, illustrated in Fig.~\ref{fig:scp}.
Sensory commutativity occurs when two sequences of actions played in different orders lead to the same final sensory state. In order to study the commutation properties of sequences of actions, we introduce Sensory Commutativity experiments (SC-experiments), which consists in having the agent play an action sequence in two different orders from the same starting point. Sensory Commutativity Probability (SCP) of a degree of freedom is then a measure of how likely two sequences of actions (in different orders) on this degree of freedom will lead to Sensory commutation.  
Note that learning SCP is a priori dependent on the environment and the morphology of the agent.  

We show that this value has intrinsic meaning for the embodied agent: if the SCP is high then the degree-of-freedom has a low impact on the environment (e.g. moving a shoulder is more likely to lead to environment changes than moving a finger, so SCP for a shoulder is lower than for a finger). By computing the SCP for each degree of freedom of the agent, we are able to characterize its motor space without any a priori knowledge and use this information for subsequent tasks. In our experiments, we illustrate how SCP, and more generally SC-experiments, can be used to learn about objects in the environment and improve sample-efficiency in a Reinforcement Learning (RL) problem. 
Our contributions are therefore the following:

\begin{itemize}
    \item We provide a mathematical framework to express sensory commutativity, and theoretical insights on how it can be useful for building perception for an embodied artificial agent.
    \item We introduce Sensory Commutativity experiments and the Sensory Commutativity Probability criterion: tools based on the commutative properties of action sequences that allow learning about the agent and the environment.
    \item We provide methods to compute them, including in realistic robotics setups.
    \item We experimentally show how SC-experiments and SCP can be useful for object discovery and improving sample-efficiency in a RL setup. Our code is available in the supplementary material.
\end{itemize}

\section{Related work and motivation}

\subsection{Related work}

SensoriMotor theory (SMT) is a theory of perception that gives prominence to the role of motor information in the emergence of perceptive capabilities \cite{o2001sensorimotor}. Inspired by philosophical ideas formulated more than a century ago by H.Poincaré \cite{poincare1895espace}, it led to theoretical results regarding the extraction of the dimension of space \cite{laflaquiere2012non}, the characterization of displacements as compensable sensory variations \cite{terekhov2016space}, the grounding of the concept of point of view in the motor space \cite{laflaquiere2013learning, laflaquiere2015learning}, as well as the characterization of the metric structure of space via sensorimotor invariants \cite{laflaquiere2018discovering}. The present work studies the commutativity of action sequences with respect to sensory information and takes inspiration from this literature.

An important insight from this literature is that action and sensor spaces have a shared underlying structure, since they are causally linked (sensory changes are caused by actions). It is suggested that the group structure would be well adapted to describe those links \cite{philipona2008developpement, poincare1895espace}, yet it has never been formalized in these works. However recently, Symmetry-Based Disentangled Representation Learning (SBDRL) \cite{higgins2018towards, caselles2019symmetry} used group theory to formalize disentanglement in Representation Learning using symmetries, i.e. transformations of the environment that leave some aspects of it unchanged. Groups are composed of these transformations, and group actions are the effect of the transformations on the state of the world and representation. Inspired by this approach, we formalize the group structure suggested in the SMT theory and use it to define the SCP criterion.

More generally, the idea of learning how actions influence sensations, and how this information can be used for exploration has been investigated in many ways. A large body of work has investigated developmental robotics \cite{schmidhuber1991curious, oudeyer2007intrinsic}, with for instance a concept related to the present work called the slowness principle \cite{luciw2013intrinsic}. 
The idea is that meaningful sensory dimensions change slowly even in the case of rapid actuator changes, which allows identifying meaningful structures such as objects. 
With the SCP criterion, we actively apply action sequences in different orders and observe the difference in sensors in order to organize useful degrees of freedom of the agent in terms of how much they impact sensors. This general idea is also present in the psychology and neurosciences literature, and is termed proximo-distal principle \cite{stulp2018proximodistal}: the tendency in infants for more general functions of limbs to develop before more specific or fine motor skills. 
This principle is also visible with the SCP, which allows to explore sensorimotor relations by prioritizing degrees of freedom which lead to bigger sensory changes: fine motor skills have high SCP and general function of limbs have low SCP.

These principles can be applied to acquire meaningful state representations in order to learn how to act in the environment. Our main motivation is to give insights on how sensory commutativity can allow seeing the problem in a novel way. We investigate two applications problems: object detection and sample-efficiency in Reinforcement Learning (RL). For object detection, we either have well-performing methods based on computer vision algorithms and largely annotated databases \cite{he2017mask}, or algorithms based on data collected by the agent itself \cite{craye2015exploration, lyubova2015, jonschkowski2019towards}. With sensory commutativity, we fall in the second category, as we aim at using sensory commutativity as the tool for detecting objects that the agent can interact with. About sample effiency in RL, the problem is often dependent on representations that are used as states. Most recent solution aim at improving the decision making component of the problem by building a new learning algorithm (HER \cite{andrychowicz2017hindsight}, SAC \cite{haarnoja2018soft}, PPO \cite{schulman2017proximal}, and many more) which are comparatively better on standard benchmarks. Here, we do not improve the learning algorithm, but rather try to show that by knowing the agent better (by computing its SCP criterion), we can improve sample-efficiency in RL by modifying its exploration strategy.

\subsection{Motivation}

Poincaré \cite{poincare1895espace} suggested that the set of compensable transformations of the environment together with the composition operation forms a group, while \cite{philipona2008developpement} further attempted at describing this group. Using action sequences and their commutative property, the authors suggested that spatial transformations and non-spatial transformations can be disentangled. 

In this paper we build on those previous works by considering the set of action sequences, termed $Seq(\mathcal{M})$, and their commutative properties. We study the group and sub-group properties of $Seq(\mathcal{M})$, with the aim of organizing the motor space $\mathcal{M}$ hierarchically. This will be achieved with the definition of the Sensory Commutativity Probability criterion.


\section{Commutative properties of action sequences}
\label{sec:commu}

\subsection{Formalism choice}

In the SMT theory, the agent sensory motor experience is described as follows:

\begin{equation}
\label{eq1}
    s_{t} = \phi (m_t, \epsilon_t)
\end{equation}

This formalism, while close to the RL formalism, is centered around the agent and its perception. At a time $t$, the agent is in a particular motor state $m_t$. 
This means that its motors are in a particular setup called $m_t$ (e.g. the actuator' torque and angle). 
The environment is defined by everything that's not the agent. It's thus an entity that is in a state $\epsilon_t$, e.g. a room with 6 walls plus light sources and objects placed in different locations. The agent can perceive the world through its sensorimotor dependencies $\phi$: a function that takes as input $m_t$ and $\epsilon_t$ and produces sensory inputs from its sensors $s_t$. 

Next, we would like to describe the dynamics of the world. This description is generally not present in SMT theory. Thus Eq.\ref{eq1} is not sufficient to support the description of the dynamics of the world. We propose to model these dynamics with the following equation:

\begin{equation}
\label{eq2}
    m_{t+1}, \epsilon_{t+1} = f(m_t, \epsilon_t, \Delta_{m_{t}}^{m'_{t+1}}, \Delta_{\epsilon_{t}}^{\epsilon'_{t+1}})
\end{equation}

The agent can operate motor commands $\Delta_{m_{t}}^{m'_{t+1}}$, which will in turn change it's sensory inputs to $s_{t+1}$ through the function $\phi$. The environment can change also and influence the agent, represented by $\Delta_{\epsilon_{t}}^{\epsilon'_{t+1}}$. Taking the initial states and changes as inputs, the function $f$ yields the new motor command $m'_{t+1}$, and a new configuration of the environment $\epsilon'_{t+1}$. We don't generally have that $\epsilon_{t+1} =\epsilon'_{t+1}$ or $m_{t+1} = m'_{t+1}$ since the agent can affect the environment configuration through its motor commands or the environment can force movements on the agent.


In summary, by combining Eq\ref{eq1} and Eq.\ref{eq2}, we obtain an equation that includes the dynamics of the world in classical SMT formulation:

$$s_{t+1} = \phi (m_{t+1}, \epsilon_{t+1}) = \phi ( f(m_t, \epsilon_t, \Delta_{m_{t}}^{m'_{t+1}}, \Delta_{\epsilon_{t}}^{\epsilon'_{t+1}}))$$

\subsection{Group structure of the set of action sequences $Seq(\mathcal{M})$}

We will now formalize groups and sub-groups of symmetries in the case of an agent moving in its environment. We study the set of motor command (or action) sequences of finite length, referred to as $Seq(\mathcal{M})$, and will attempt at describing its structure.

\begin{figure*}[h!]
    \centering
    \includegraphics[scale=0.4]{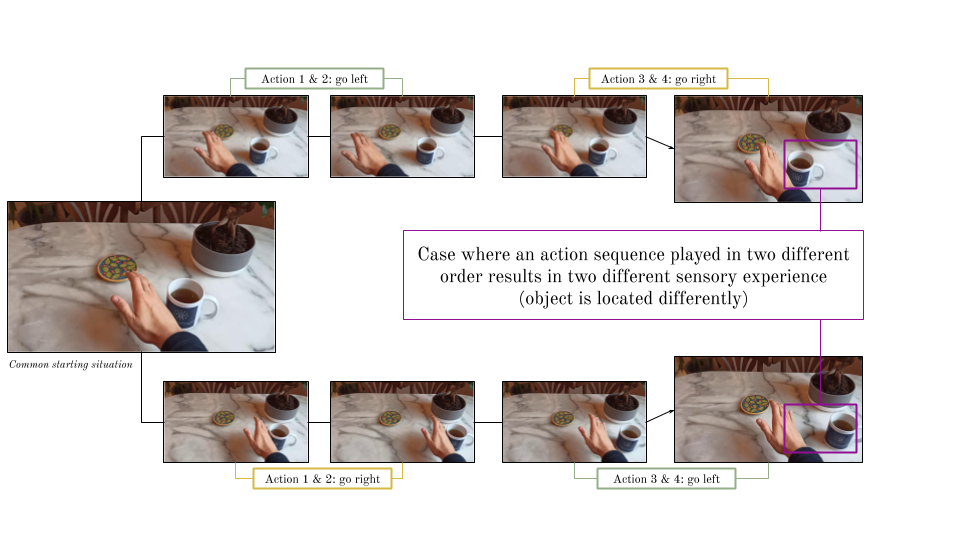}
    \caption{Example of SC-experiment that does not commute. Starting from a common situation, the action sequence played in two different orders does not lead to the same sensory state.}
    \label{fig:illu}
\end{figure*}

Philipona \cite{philipona2008developpement} first defined a relation between action sequences: $h \sim g $ if and only if $h$ and $g$ affect the sensors in the same way. Using our formalism, we can translate this concept into an equality. 
\begin{defn}
Let $(h,g)\in Seq(\mathcal{M})$. h is equivalent to g under $(m_t, \epsilon_t)$, noted $h\sim_{m_t, \epsilon_t} g$ if and only if they produce the same sensory states when applied from the same starting situation of the agent ($m_t$) and the environment ($\epsilon_t$):

$$h\sim_{m_t, \epsilon_t} g \iff \phi(f(m_t, \epsilon_t, h,\Delta_{\epsilon_{t}}^{\epsilon_{t+1}})) = \phi(f(m_t, \epsilon_t, g,\Delta_{\epsilon_{t}}^{\epsilon'_{t+1}}))$$

\end{defn}
Intuitively, two actions sequences are equivalent for a particular motor state and environment state if applying them lead to the same sensory state.
For instance in the case of multiple-joints arm moving freely in an empty space, there are multiple different ways of moving the arm from one motor state to another. This yields action sequences ($h_1, .., h_n)$ which are equivalent in this situation $(m_t, \epsilon_t)$, we thus have $h\sim_{m_t, \epsilon_t} g$. However in other situations these actions sequences can become not equivalent, for instance if there are objects on the way as illustrated in Fig.~\ref{fig:illu}.

For convenience and clarity, we will drop the notation for dependence on $(m_t, \epsilon_t)$ and thus write $h\sim g$ whenever there are no ambiguities in the context. We now consider the structure of $Seq(\mathcal{M})$ under composition $\circ$ with respect to the equivalence $\sim$.



\begin{prop}[Structure of ($Seq(\mathcal{M}$), $\sim$, $\circ$)] 
\label{prop1}
The following properties hold: \\
1. $\sim$ is an equivalence, i.e. it is reflexive, transitive and symmetric.\\
2. ($Seq(\mathcal{M})$, $\circ$) is a group w.r.t $\sim$.\\
3. $\circ$ is not commutative with respect to $\sim$.

\end{prop}
\begin{proof}
\begin{enumerate}
    \item $=$ is an equivalence, thus $\sim$ is an equivalence as well.
    \item All 4 properties of the group definition are satisfied. (i) For two action sequences $(h,g)\in Seq(\mathcal{M})$, the composition of $h$ and $g$ is still an action sequence $h\circ g \in Seq(\mathcal{M})$. (ii) $\circ$ is associative with respect to $=$, i.e. $g\circ (h \circ k) = (g \circ h) \circ k $ thus it follows that $g\circ (h \circ k) \sim (g \circ h) \circ k$. (iii) The identity element is the no-op action. (iv) If we suppose that there are no irreversible phenomenons in the environment, then for a fixed $(m_t, \epsilon_t)$, all action sequences can be inverted. 
    \item $\circ$ is not commutative, as we can always explicitly find two action sequences that do not commute. For instance once there exists a movable object in the environment: if the agent is placed left to the object, then let $h$ be moving right and $g$ be moving left. $h$ and $g$ do not commute (Fig.~\ref{fig:illu}).
\end{enumerate}
\end{proof}


($Seq(\mathcal{M})$, $\circ$) is thus a group w.r.t $\sim$. This structure is consistent with the intuitions in SBRL and SMT theories. In the following, we build on the observation that composing action sequences is not generally commutative as we can measure to which degree they commute. We show how this property can lead the agent to organize and interpret its motor space.

\subsection{Philipona's conjecture}
\label{sec:conjecture}

Philipona \cite{philipona2008developpement} already studied how action sequences commute with respect to the sensory information received by the agent. Notably, Philipona defined commutative residues. Suppose that an agent doing $h_1 \circ h_2$ leads to a different outcome in observations than doing $h_2 \circ h_1$, then a commutative residue $g$ is an action sequence that the agent has to do to compensate the difference in sensory experience.

\begin{defn}
$g$ is a commutative residue of $(h_1, h_2)$ if and only if $h_1 \circ h_2 \sim_{s} h_2 \circ h_1 \circ g$. If $g$ is equivalent to no-op (no action), then $h_1$ and $h_2$ commute.
\end{defn}

Starting from this definition, he conjectured that all action sequences that are not displacements commute with any action sequences. For instance, moving your arms (displacement action) then opening the eyes (non-displacement action) will always commute whereas two displacement actions will not necessarily commute, depending on which starting situation $s$ is selected. 

\begin{conj}[Philipona's conjecture]

\hspace{0.1cm} Let $Seq(\mathcal{M})$ be the set of action sequences. The subset of $Seq(\mathcal{M})$ composed of non-displacements action sequences is the sub-group of $Seq(\mathcal{M})$ that commutes.

\end{conj}





We will illustrate this conjecture with experiments in Sec.~\ref{sec:conjecturephilipona_exp}.

\subsection{Sensory commutativity probability of an action sequence}

Based on Philipona's conjecture, we derive a criterion for characterizing how much each degree of freedom of the agent affects the world, computable using only sensorimotor data. We define "degree of freedom" (DOF) as a dimension of the multidimensional continuous action space of the agent. We also define what we term a sensory commutativity experiment: for an action sequence $h$, the agent plays it in two different orders starting from the same situation.

\begin{defn}[Sensory commutativity experiment (SC-experiment)]

Let $h$ be an action sequence of finite length. Let $h_p$ be a random permutation of $h$ (same sequence but different order). 

We define a sensory commutativity experiment (SC-experiment) as playing $h$ and $h_p$ from the same starting point and comparing the two resulting observations in the agent's sensors.

\end{defn}

Using the conjecture, we have that for an SC-experiment, the agent can experience two different sensory outcomes only if the action sequence $h$ is composed of at least one displacement action (an action that affects the environment such as moving limbs or going forward). 

However, not all displacement actions are equivalent. The agent is more likely to observe two different outcomes if the action sequence is composed of displacement actions that affect the environment \textit{a lot}. Consider moving your forearm (elbow joint) compared to moving your whole arm (shoulder joint): the latter is more likely to move things around in the environment and thus induce sensory non-commutativity when played in two different orders (i.e. having two different sensory outcomes). An elbow joint should therefore have a higher SCP than a shoulder joint. 

We formalize this intuition by defining the Sensory Commutativity Probability (SCP) of a degree of freedom, averaged over all starting situations~$s$: 

\begin{defn}[Sensory commutativity probability of a degree of freedom]

Let $Seq(\mathcal{M}_k)$ be the set of motor commands (or action) sequences of finite length for the k$^{th}$ degree of freedom of $\mathcal{M}$ (motor state space). Let $h\in Seq(\mathcal{M}_k)$ and let $h_p$ be a random permutation of $h$ (same sequence but different order).

The Sensory Commutativity Probability of the k$^{th}$ degree of freedom $SCP(\mathcal{M}_k)$ is defined as: 

$$SCP(\mathcal{M}_k) = \mathbb{P}_{s, h}[h\sim_{s} h_p]$$

\end{defn}

\subsection{Sensory Commutativity Probability computation}
\label{sec:scp_computation}

We propose a straightforward procedure to estimate the SCP of each degree of freedom of the agent. We initialize the SCP value to $0$ (\texttt{SCP$\leftarrow$0}). We then repeat the following process $n$ times for each DOF:

\quad - Sample an action sequence using the selected degree of freedom (a sequence of action where each action is a value between -1 and 1).

\quad - Play it in 2 different orders starting from the same randomly chosen state and save the two final sensor images $s_1$ and $s_2$. Compute the distance between the two images $d(s_1, s_2)$. 

\quad - Count one (\texttt{SCP+=1}) if $d(s_1, s_2) \leq t$, zero otherwise. 

Finally, the estimator of the SCP is the average over the number of trials (\texttt{SCP$\leftarrow$SCP/$n$}). 

The parameters of the algorithm are the selected distance function $d$ that allows comparing the agent's observations, the threshold $\tau$, and the number of iterations $n$. Note that using a simulation allows playing the two action sequences of different orders from the exact same starting position. We discuss the need for simulation to compute SCP and more generally SC-experiments in Sec.~\ref{sec:discu} and how to overcome this requirement for real-life experiments.

\subsection{SC-experiments for object detection}
\label{sec:obj_theory}
The concept of SCP is based upon comparing outcomes of SC-experiments and evaluating whether the two resulting observations are considered equal or not. Going beyond this equality test, we propose to have a finer analysis of the differences between the two observations $obs_{1}$ and $obs_{2}$ resulting from an SC-experiment.

Comparing $obs_{1}$ and $obs_{2}$ leads to three possible outcomes from which the agent can learn about immovable and movable objects in the environment. 

\begin{itemize}
    \item $obs_{1}$ and $obs_{2}$ are entirely different: the two action sequences from this starting position do not commute, because the agent interacted with immovable objects. Using the position of the agent, we can now map immovable objects in the environment.
    \item $obs_{1}$ and $obs_{2}$ are identical: the two action sequences from this starting position commute, because the agent did not interact with anything in the environment (free movement). Using the position of the agent, we know that there are no objects in the current space around it.
    \item $obs_{1}$ and $obs_{2}$ are identical except for a limited area corresponding to an object that has been moved: it's the case where the agent has interacted with a movable object that did not block the agent's movement. Hence the two action sequences would have commuted for most of the environment, except for the object that has been moved. We can learn to detect this moving object and track it.
\end{itemize}

\subsection{Experiments}

In order to illustrate all these concepts, the experiments presented in the remainder of this paper are organized as follows: we first show how to compute SCP in 2D simple environments, then in 3D realistic robotic setups. Then, we show how we can use SC-experiments to learn about immovable and movable objects in realistic robotics setups. Finally, we show how SCP can be used for improving sample-efficiency in RL. Our code is attached in the supplementary material.

\section{Sensory Commutativity Probability experimental analysis}
\label{sec:scpp}

In this first experimental section, we compute and interpret the SCP in a 2D and a 3D embodied agent scenarios. 
In order to study the properties of SCP and how it relates to the emergence of the notion of objects, we use simulation environments that have the following properties: embodied agent, navigable space with objects to interact with, first-person high-dimensional observations, low-level high-dimensional action space, and coherent physics.

\subsection{2D experimental setup}


\begin{figure}[h]
\centering
\includegraphics[scale=0.23]{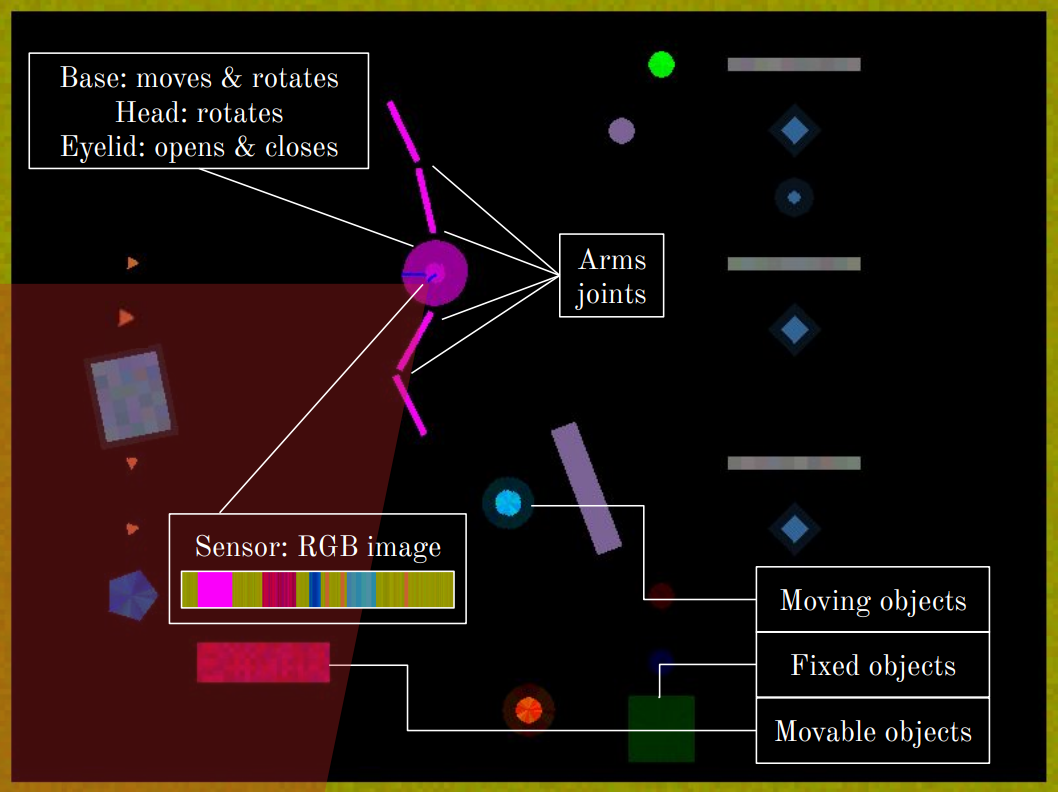}
    \caption{Simulation used for our experiments. The agent Polyphemus has a 8 DOF motor space, receives an image of it's only eye, and is placed in a room with fixed, movable and moving elements.}
    \label{fig:sm_simu}
\end{figure}

\quad \textbf{Simulation description.} 
Our first experiment uses Flatland \cite{caselles2018flatland}, a platform for creating 2D RL environments. We construct an agent called Polyphemus (a Cyclop from the Greek mythology), that has a base that can move forward and rotate, a rotatable head and two 2-DOF arms. The agent sees through its unique eye that has an activable eyelid, for a total of 8 DOF. The observation received by the agent is a 64x3 line of RGB pixels (as the world is 2D), which corresponds to the field of view of 90 degrees. 
This agent is placed in a room with fixed, moving, or movable entities, all of different colors. It can move around and physically interact with these entities. Its point of view can change through base movement, rotation, and head rotation. Our simulation is illustrated in Fig.~\ref{fig:sm_simu}. For each degree of freedom, an action or motor command corresponds to a change in the longitudinal/angular velocity of the degree of freedom.



\begin{figure}[h]
    \centering
    \includegraphics[scale=0.18]{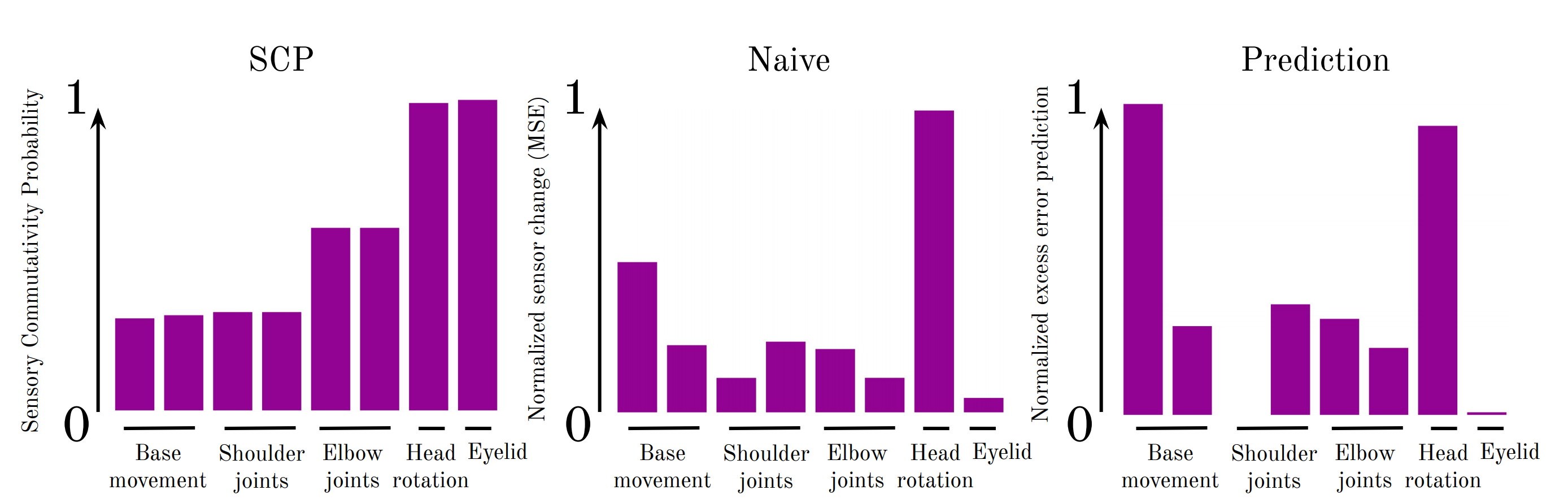}
    \caption{\textbf{Left:} Sensory Commutativity Probability for each degree of freedom. \textbf{Middle:} Naive alternative. \textbf{Right:} Prediction error  alternative.}
    \label{fig:conj_res_exp}
\end{figure}

\quad \textbf{SCP computation.} In order to compute the SCP of each of the 8 agent's degrees of freedom (Fig.~\ref{fig:conj_res_exp}, left), we have to select a distance and threshold as mentioned in Sec.~\ref{sec:scp_computation}. The distance selected here is simply the mean squared error between $s_1$ and $s_2$, the observations resulting from the two sequences of actions of a SC experiment. Because there is no noise in the dynamics of the environment and the sensor, the future of the agent is deterministic. Therefore, in this particular case we can use a threshold of $0$. This means that we consider that two action sequences sensory commutes if and only if applying the two action sequences from the same initial state lead to exactly the same sensors. This hard constrain will be relaxed in subsequent experiments (Sec.~\ref{sec:3Dsetup}).

\quad \textbf{Baselines.} The SCP criterion derived in this paper estimates how much each degree of freedom affects the environment in an embodied agent scenario. We tried two alternatives to this approach in order to estimate the same quantity. A straightforward approach to this problem, which we call the naive alternative (Fig.~\ref{fig:conj_res_exp}, middle), is to play action sequences of each degree of freedom and quantify how much the sensors change. A more involved approach is to use prediction on the sensory change caused by each degree of freedom (Fig.~\ref{fig:conj_res_exp}, right), a common approach used to improve exploration in RL \cite{burda2018exploration, pathak2017curiosity}. We call this alternative the prediction error approach. The DOF that are harder to predict could be the ones affecting the environment the most, and thus the most important for manipulation and navigation.

\subsection{3D realistic experimental setup}
\label{sec:3Dsetup}
We also compute and interpret the SCP for a realistic embodied agent scenario using the interactive Gibson environment (iGibson) \cite{xia2020interactive}.

\begin{figure*}[h]
    \centering
    \includegraphics[scale=0.3]{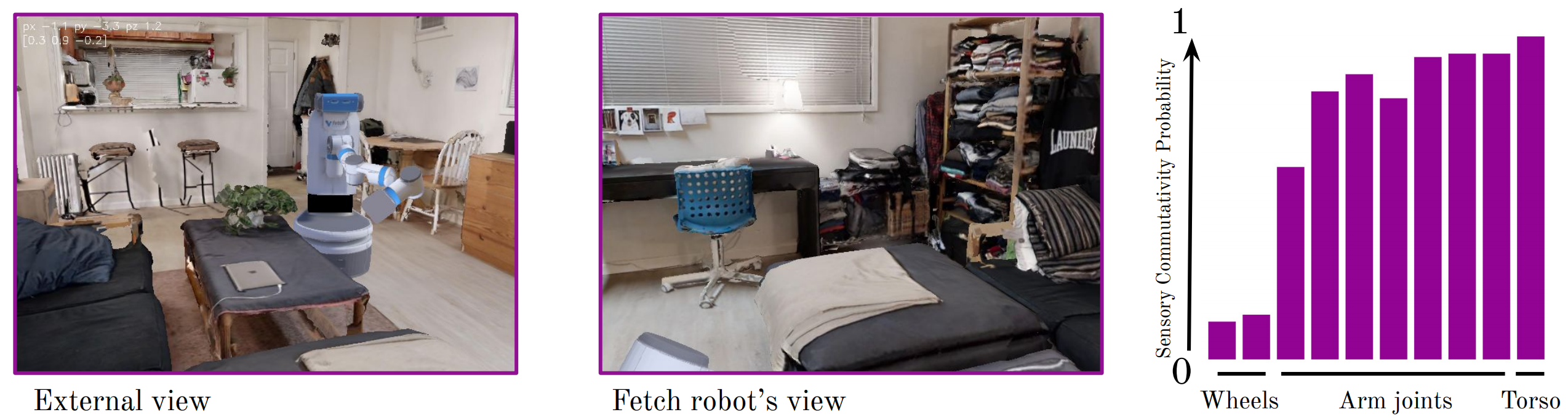}
    \caption{\textbf{Left:} External view of the iGibson simulator where the Fetch robot is in a living room. \textbf{Middle:} Fetch's first person view. \textbf{Right:} SCP computed for each of Fetch's degrees of freedom.}
    \label{fig:igibson}
\end{figure*}

\quad \textbf{Simulation description.} iGibson is a simulation environment for robotics providing fast visual rendering and physics simulation. It is packed with a dataset with hundreds of large 3D environments reconstructed from real homes and offices, and interactive objects that can be pushed and actuated. In our experiments, we use the Rs environment, which is basically a regular apartment. We place the Fetch robot in this environment (Fig.~\ref{fig:igibson}, left). Fetch is originally a 10-DOF real robot \cite{wise2016fetch} equipped with a 7-DOF articulated arm, a base with two wheels, and a liftable torso. Fetch perceives the environment through a camera placed in his head (Fig.~\ref{fig:igibson}, middle). 

\quad \textbf{SCP computation.} In the Flatland environment, two action sequences commuted only if the sensory result of applying both from the same starting situation was perfectly equal. We relax the strict equality condition to compute the SCP for Fetch (Fig.~\ref{fig:igibson}, right). Indeed, with real images, only an offset of one pixel would render the two action sequences non-sensory commutative. Instead of using the mean squared error as a distance, we use a perceptual distance using the VGG16 \cite{simonyan2014very} features of each observation. We thus have $d(s_1, s_2) = ||VGG16(s_1) - VGG16(s_2)||^2_2$. The choice of the threshold $\tau$ is partly arbitrary, as we are interested in relative comparisons between degrees of freedom. We verify in our experiments that our results and conclusions are valid for a large range of $\tau$.

\subsection{Results}
\label{sec:conjecturephilipona_exp}

\quad 
In the Flatland environment, Fig.~\ref{fig:conj_res_exp} (Left) shows that only two actions have an SCP of 1: \textit{eyelid} and \textit{head rotation}. All other actions have an SCP inferior to 1. \textbf{This is consistent with Philipona's conjecture} (Sec.~\ref{sec:conjecture}): \textit{eyelid} and \textit{head rotation} are the two degrees of freedom that are not associated with displacements, thus action sequences composed of actions of these type commute with respect to the sensors. On the contrary, all other degrees of freedom are associated to displacements, and thus will eventually induce non-zero commutation residues when played in different orders from the same starting situation. We observe the same results in iGibson, presented in Fig.~\ref{fig:igibson}: the torso lift DOF is not associated with displacement in the environment, so it has an SCP of 1, i.e. it always sensory commutes. Hence the results are consistent with the conjecture and can be used by the agent to autonomously discover which of its actions are associated with displacements or not. 

\quad \textbf{Qualitatively, SCP is inversely proportional to how each degree of freedom affects the environment.} By that we mean that from the computation of the SCP, we obtain a hierarchical organization of the action space in which the more important dimensions for manipulation and navigation are separated from the dimensions that are not crucial for such tasks. For instance, we inferred that shoulders should have a lower SCP than elbows since activating the shoulder joint is more likely to induce non-commutativity by moving things around or hitting walls/obstacles. This intuition is verified by our results. Shoulders and base movement have a lower SCP than elbows which in turn have a lower SCP than eyelid and head rotation, as observed in Fig.~\ref{fig:conj_res_exp}. Without having any prior knowledge about the simulation, we can automatically organize the agent's degrees of freedom in a hierarchy. Moreover, the symmetry of the action space is kept, as elbow 1 and 2 have equal SCP, and so do shoulder 1 and 2. We reach the same conclusions on iGibson (see Fig.~\ref{fig:igibson}, right). The wheels have the lowest SCP since they provide longitudinal movement and rotations for the robot. Then comes the first DOF of the articulated arm, i.e. the ones that are closer to its base (like shoulders vs. elbows in the Flatland experiments). Finally, the highest SCP values  correspond to the arm DOF that are further on its arm and the torso lift. Once again, we obtain a hierarchical organization of the action space in which the less important dimensions for manipulation and navigation are separated from the dimensions that are not crucial for such tasks.

About the choice of the threshold to compute the SCP, we tried a range of values for $\tau$, from $20$ to $100$, and in each case, we obtain the same hierarchy and thus the same conclusion, only the nominal values change, which is irrelevant for the use of SCP.

In additional experiments presented in \ref{app:add_exps}, we verified the robustness of these results. We computed the SCP for 8 different combinations of agents and environments (longer/smaller arms, more/fewer objects) and confirmed our intuitions on the interpretation of SCP described above. In additional experiments presented in \ref{app:add_exp_igibson}, we also verified the robustness of these results in iGibson by computing the SCP for a different type of robot called JackRabbot \cite{martin2019jrdb}. We reach the same conclusions as with the Fetch robot.

\quad \textbf{Alternative methods are not adapted.} Details for these two experiments are available in \ref{app:alternatives} and results are illustrated in Fig.~\ref{fig:conj_res_exp}. Both approaches fail to replace the SCP criterion. We see that for the naive approach, rotating the head of the agent changes dramatically what the agent sees, even though this degree of freedom does not affect the environment. For the prediction error alternative, we see the same problem with head rotation and a great difference between the two base movements (rotation and longitudinal movement) while they affect the environment in similar ways. Indeed, it's harder to predict what's outside the field of view of the agent so rotation is harder to predict compared to longitudinal movement. To conclude, the proposed alternatives could not yield the same organization of the agent's DOF.

\section{Applications of Sensory Commutativity}

\subsection{Sensory Commutativity Probability for object detection}

We would like to verify the intuition described in Sec.~\ref{sec:obj_theory}:  there are three possible outcomes to an SC-experiment (different observations, identical observations, and identical observations up to moved objects) and from these outcomes, the robot can detect and map immovable and movable objects in the environment, by doing SC-experiments (playing action sequences in different orders from the same starting point and comparing the resulting observations $obs_{1}$ and $obs_{2}$). Our experiments are performed in iGibson with the Fetch robot.

\begin{figure*}[h!]
    \centering
    \includegraphics[scale=0.28]{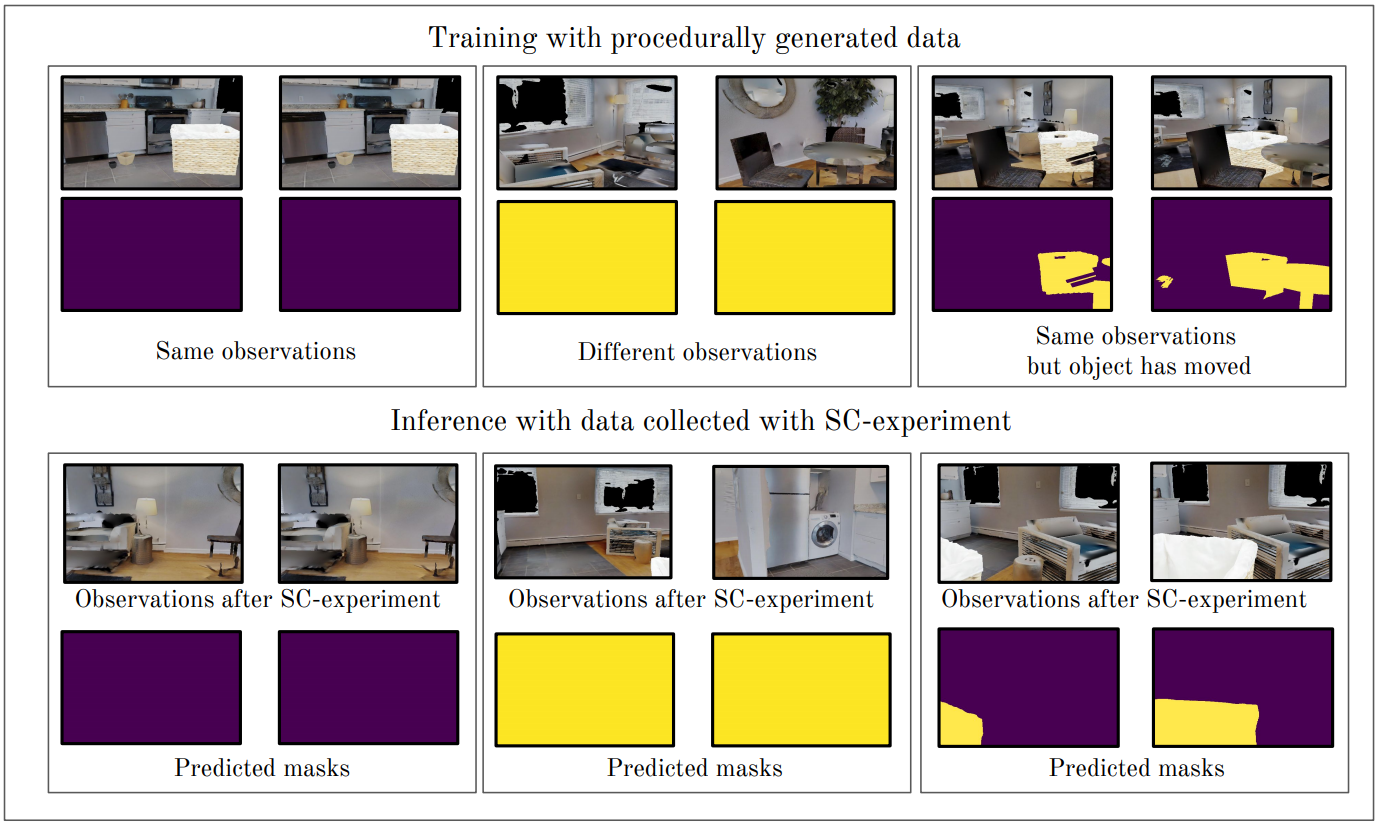}
    \caption{Dataset for training the mask predictor and inference results on data collected with SC-experiments. The dataset is procedurally generated to simulate the three possible scenarii resulting from a SC-experiment. \textbf{Left:} scenario where there are no changes in the observations. \textbf{Middle:} scenario where the observations are different. \textbf{Right:} scenario where the observations are identical up to moved objects.}
    \label{fig:dataset_obj}
\end{figure*}

\subsubsection{Method}

In order to verify the aforementioned intuition, Fetch needs to be able to perform an SC-experiment and then detect: 1) if the two resulting observations are identical or not, 2) if they are identical except for the parts of an image corresponding to an object that moved. Studies in cognitive science indicate that children are capable of doing this differentiation at a very young age (1 month old) \cite{kaufmann1995development, johnson2010infants}, so we consider that equipping the agent with this basic ability is a reasonable assumption. Therefore, we equip the agent with a vision system that gets two observations as input and outputs two masks which will be all zeros if the two observations are identical, all ones if they are different, and the mask of the modified area if something has changed.

We thus train a neural network with generated data to predict those two masks with two observations as input. We refer to this model as the "mask predictor". 

\quad \textbf{Dataset.} The data is collected in the Placida environment by starting at a random position in the environment (observation $obs_1$) and then collecting data for the three possible outcomes:
\begin{itemize}
\item no difference: it suffices to keep the same observation and the corresponding masks are all zeros. The data is ($obs_1$ + all zeros mask, $obs_1$ + all zeros mask).
\item completely different: we move the robot and get a different observation $obs_2$, the corresponding masks are all ones. The data is ($obs_1$ + all ones mask, $obs_2$ + all ones mask).
\item no difference except moved objects: we randomly disturb the orientation and position of some movable objects and get a new observation $obs_2$ identical to $obs_1$ up the moved objects. The data is ($obs_1$ + moving objects mask, $obs_2$ + moving objects mask)
\end{itemize}

The resulting dataset is illustrated in Fig.~\ref{fig:dataset_obj}. The objects we use for training are the original objects found in the interactive Placida environment, augmented with several object from the YCB object benchmark \cite{calli2015benchmarking}.

\quad \textbf{Architecture and training.} We then train the neural network to predict the masks given the observations. This process is similar to predicting the optical flow of two consecutive frames in a video. 
Thus, for the mask predictor, we compared FlowNet-S \cite{fischer2015flownet}, a popular baseline for optical flow prediction, with the state-of-the-art RAFT model \cite{teed2020raft}, and selected RAFT because of its higher prediction accuracy. 
We train the model using the same architecture and optimization process as in the paper, except for the loss function and the output activation function. We change the loss function to a binary cross-entropy loss between the ground truth mask and the output mask of the network. We select the sigmoid function as output activation function so that the model outputs binary masks instead of the original optical flow map output ($2*W*H$). All training details are available in the original open-source implementation we used\footnote{\url{https://github.com/princeton-vl/RAFT}}.

\quad \textbf{Inference.} Once the mask predictor is trained, we place the agent in an environment and perform SC-experiments where we let it play an action sequence in different orders from the same starting point. Then, the goal is for the agent to detect immovable and movable objects using the generated data from the SC-experiments and the mask predictor. All experimental details are described in \ref{app:exp_obj}.

\quad \textbf{Evaluation.} For qualitative and quantitative evaluation, we manually create a test set with $50$ tuples $(obs_1, obs_2, mask_1, mask_2)$ of the three possible scenario resulting from a SC-experiment. We cannot construct this dataset automatically, as the mask has to be manually created by either assessing if the two observations are different or identifying which object has moved between the two observations. Using this dataset, we can first assess the prediction accuracy among the three possible scenarios. 

In the case where an object has moved (see example in lower right corner of Fig.~\ref{fig:dataset_obj}), we can further analyze the accuracy of the predicted mask using the Jaccard index, or Intersection over Union ($IoU$). It quantifies the overlap between predicted $(p1, p2)$ and ground-truth $(gt_1, gt_2)$ masks. It is defined as $IoU_i = \frac{|p_i \cap gt_i|}{| p_i \cup gt_i|}$.

\subsubsection{Experiments and results}

Quantitatively, the performance of the mask predictor on the manually collected test set reach a prediction accuracy among the three possible scenarios of $82\%$. We reach an average Jaccard index of $0.85$ on the subset of instances where an object has moved (see example in lower right corner of Fig.~\ref{fig:dataset_obj}).

\quad \textbf{Do SC-type experiments allow detecting movable objects?} We compute SC-experiments using a DOF selected using SCP value. We select the DOF with SCP closest to $0.5$ in order for the outcome of SC-experiments to be as diverse as possible, i.e. the DOF of the arm that is closest to the body of the agent. 

Results presented in Fig.~\ref{fig:dataset_obj} \& \ref{fig:obj_more} show that using the mask detector with the outcome of these SC-experiments allows to detect objects that have been moved. Note that the mask detector only detects objects that have moved between the two resulting observations, rightfully ignoring the other potential objects that were not moved. After this detection, we can then use semi-supervised tracking algorithms such as \cite{oh2019video} in order to track the detected object.

\begin{figure}[h!]
    \centering
    \includegraphics[scale=0.3]{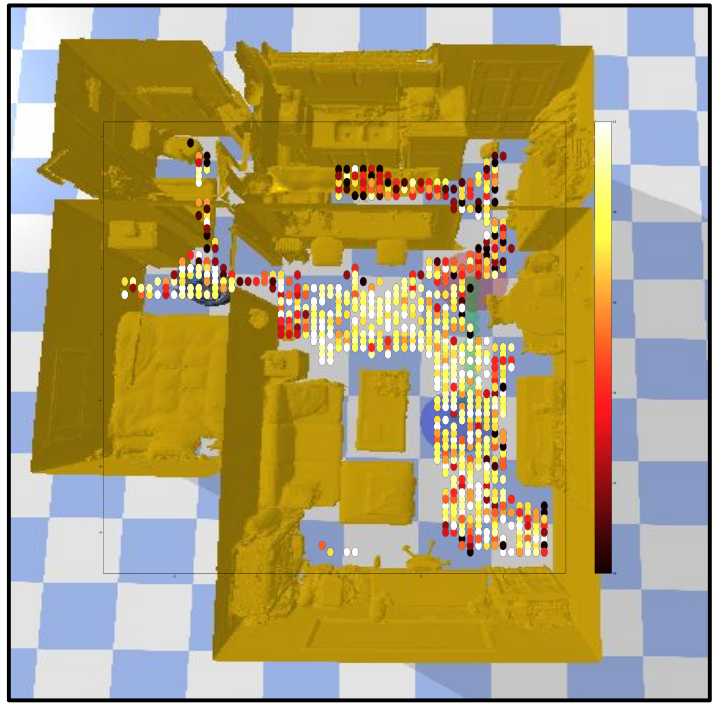}
    \caption{Local SCP value corresponding to the arm's DOF that is closer to the body of the agent, computed over 10 SC-experiments for each position in the Rs-interactive environment.}
    \label{fig:scp_map}
\end{figure}

\quad \textbf{Do SC-type experiments allow detecting immovable objects?} Results presented in Fig.~\ref{fig:dataset_obj} show that the mask predictor is also able to accurately predict when the observations are different or identical. By isolating those two cases from the case where only one or a few objects have moved, we can compute a local SCP value that tells us whether the agent interacted with an immovable object during the SC-experiment. We can compute this local SCP value for different starting positions in the environment, and then construct a map of immovable objects in the environment. We present this map in Fig.~\ref{fig:scp_map} for the arm's DOF that is closer to the body of the agent (we choose this DOF with the same reasoning as the previous result). Results show that regions with low local SCP value correspond to regions where there are walls and immovable objects in the way of Fetch's arm. 

Indeed, in the kitchen part (room at the top), the space is cramped and so most of the positions indicate low SCP (less than $0.4$) because of the interactions induced with the furniture. In the living room (main room) and the bedroom (at the left), most empty space show high local SCP (around $0.8$ and $1.0$). Notice how the local SCP value is also high around objects that are low and thus have to low chance to interact with the arm (bed in the bedroom, low table in the living room). We thus obtain a mapping of immovable objects in the environment using SCP.

\subsubsection{Generalization study} 

In principle, this movable and immovable object detection method is designed to work in any environment, any objects and any field of view. Indeed, it only relies on having a precise mask predictor, which we show can be achieved. We thus performed a generalization study of our method. We performed inference on data with objects, environments and field of view that were not shown during training. For this study, we selected the Rs-environment and the Bolton environment, objects from the YCB benchmark that were not shown during training, and a bigger field of view (90 versus 45 for training). 

\begin{figure*}[h!]
    \centering
    \includegraphics[scale=0.27]{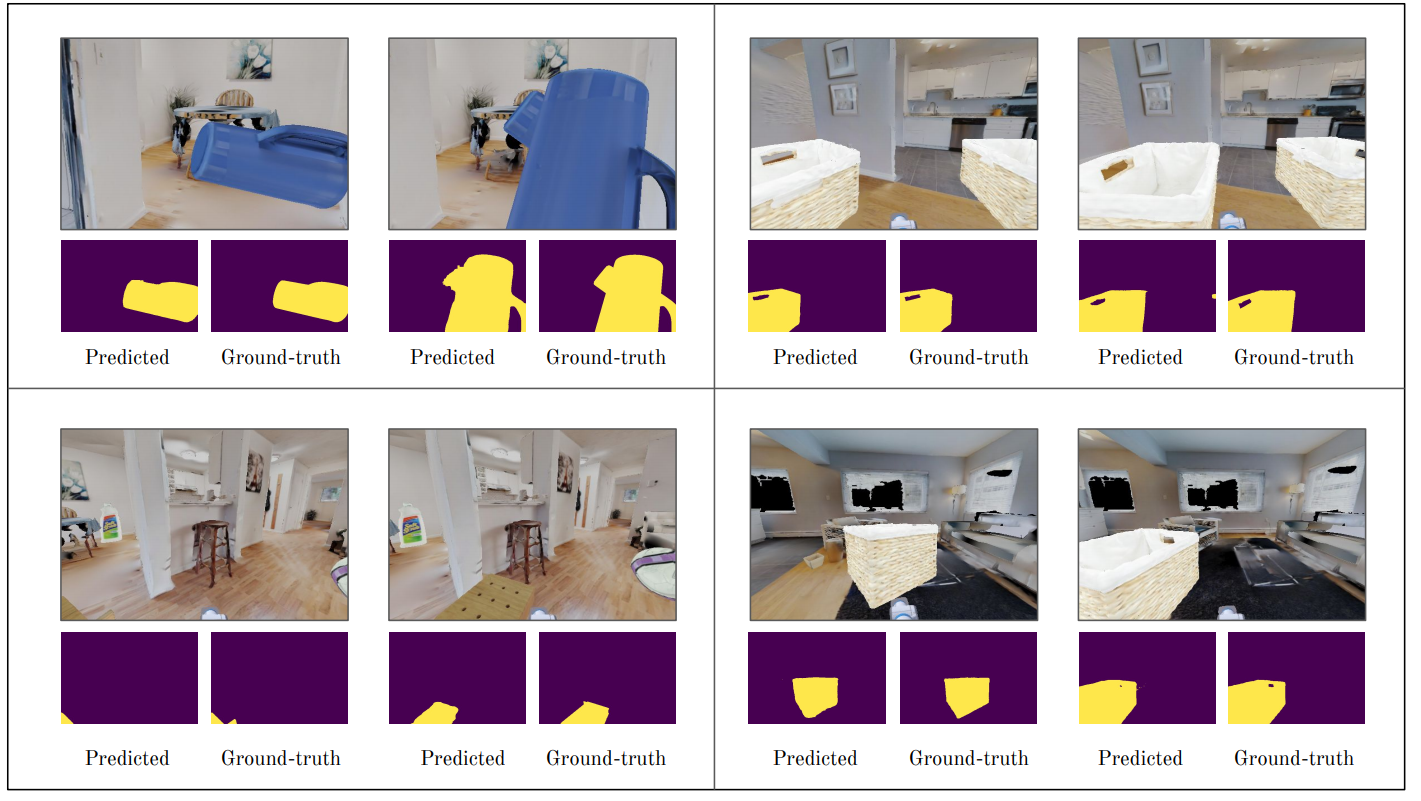}
    \caption{Generalization study of movable object detection using SC-experiments and a mask predictor trained in the Placida environment. In all scenarios, our method correctly predicts the mask object. \textbf{Upper left:} Object and environment not seen during training. \textbf{Lower left:} Object, environment and field of view not seen during training. \textbf{Upper and lower right:} Field of view not seen during training.}
    \label{fig:obj_more}
\end{figure*}

In Fig.~\ref{fig:obj_more}, we show results for the generalization study, which indicate that the mask predictor can indeed be used with environments, objects, and field of view that have been not shown during training. Qualitatively, the mask predictor seem to be able to precisely predicts which objects has moved. Quantitatively, the precision of our object detection method in those generalization scenarios is mostly not affected. We manually created another test set of 20 instances with out-of-distribution environments and objects, for the scenario where an object has moved. We reach an average Jaccard index of $0.78$, with most instances with a Jaccard index of $1$, and a few where the detection is totally missed, thus lowering the average. We observe that when the detection does not totally miss the object, the precision of the mask is excellent. The low performance drop between in-distribution and out-of-distribution test set ($0.05$) allows us to conclude that our method generalizes to new environments, objects and field of view.

\subsubsection{Alternatives are not adapted} 

Alternatives to SC-experiments such as just playing an action sequence and comparing the first and last observations would detect much fewer objects because many experiments would result in a complete image change where the SC-experiments would highlight only a particular object.
Another alternative would be to start in a position, play an action sequence, and then go back to this starting point and compare what's changed. While this approach would be comparable for movable object detection, this would not allow detecting immovable objects.

\subsection{Sensory Commutativity for efficient RL}

We now illustrate how SCP can be used for unsupervised exploration, by using it to improve sample-efficiency in an RL setup. For computational reasons, we experiment with the Flatland simulator.

\subsubsection{Experimental setup}

\begin{figure}
\centering
\includegraphics[scale=0.17]{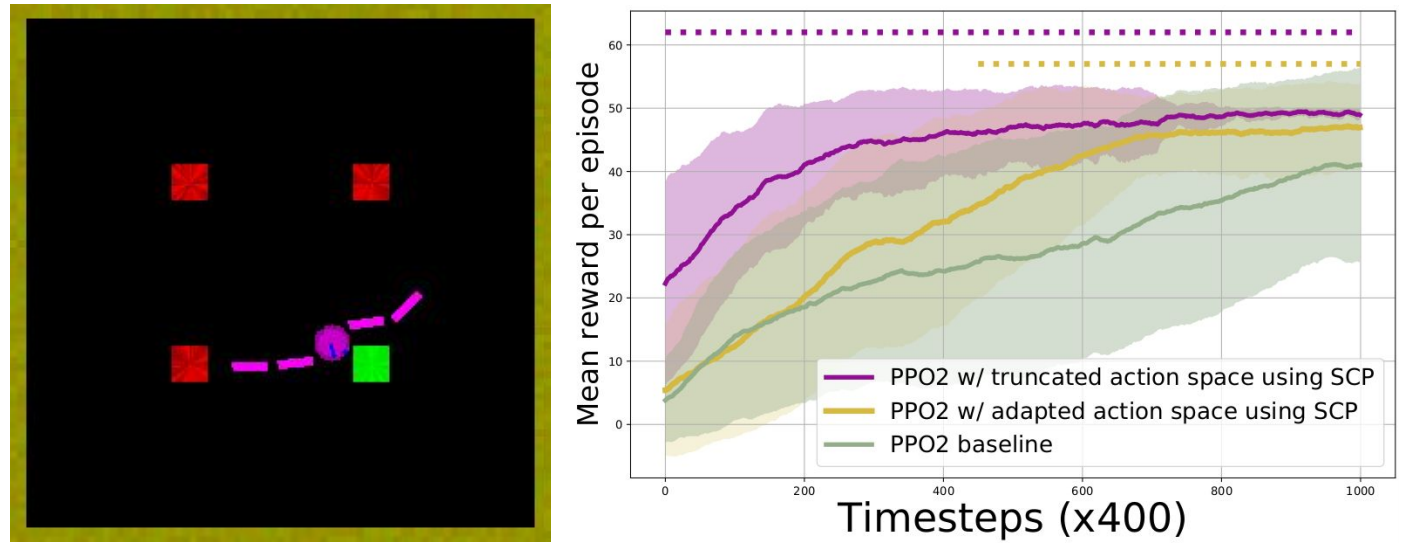}
    \caption{\textbf{Left:} RL task. \textbf{Right:} Results.}
    \label{fig:rl_exp}
\end{figure}

We use the PPO2 \cite{schulman2017proximal} implementation from Stable-Baselines \cite{hill2018stable}. The policy is composed of a 1D convolutional feature extractor followed by a recurrent policy. We consider the same agent, Polyphemus, for which we computed the SCP criterion in Fig.~\ref{fig:conj_res_exp}. The input of the policy is the RGB image of what Polyphemus' eye sees. The environment considered is a square room with 3 dead zones (which terminate the episode with a -20 reward) and a goal zone (which terminates the episode with a +50 reward), illustrated in Fig.~\ref{fig:rl_exp}. We propose two methods that take advantage of the SCP to modify the action space of the agent. The goal is to improve sample-efficiency when learning to solve a task in this embodied scenario.

\quad \textbf{SCP-truncated action space.} We propose to to focus exploration on the degrees of freedom that have a high impact on the environment, by fixating degrees of freedom corresponding to high SCP. We implement this by halving the dimension of the action space, keeping only the degrees of freedom that have the most effect on the environment, i.e. lower SCP value. We thus keep the base movement and rotation, and the shoulders joint, while discarding the elbow joints, head rotation, and eyelid activation. We refer to this method as \textit{SCP-truncated} action space. This action space reduction will simplify the RL task, as long as the necessary actions such as base motion are selected by the SCP criteria.

\quad \textbf{SCP-adapted action space.} A less involved proposition is to modify the action sampling interval according to the SCP value, for each degree of freedom. This method will modify the exploration dynamics to favor important actions. Suppose that the sampling interval for each dimension of the action space is $[-1,1]$. If a dimension has high SCP, i.e. it does not affect the environment a lot, we then reduce the interval from which actions are sampled $[-1\cdot l(SCP), 1\cdot l(SCP)]$. The function $l$ maps the highest SCP to $0$ and lowest SCP to $1$, then we use a linear interpolation between those two points to deduce values for $SCP\in ]-1,1[$. We refer to this method as \textit{SCP-adapted} action space. 

\quad \textbf{Comparison protocol.} We compare those two strategies to a baseline policy trained to solve the task with the complete action space. We average the result of each policy over 30 trials initialized with different random seeds, and we test the statistical significance of our results according to the guidelines provided by \cite{colas2018many}.

\subsubsection{Results}

The results are displayed on Fig.~\ref{fig:rl_exp}. First, we notice that all strategies are viable to solve the task. We now compare sample-efficiency between the strategies. The policy trained with \textit{SCP-truncated} action space can learn how to solve the task more than twice as fast as the baseline policy. The discarded degrees of freedom are not crucial in this navigation task, hence the agent is still able to solve the task using only the degrees of freedom that have the lowest SCP value. The policy trained with \textit{SCP-adapted} action space is less sample-effective than the \textit{SCP-truncated} but still learns significantly faster than the baseline policy.

\section{Discussion and conclusion}
\label{sec:discu}

\quad \textbf{Discussion: extending SCP and SC-experiments to real life.} The difficulty for SCP and SC-experiments in real-life is that the agent has to be able to play two action sequences from the same starting point. Thus, in a real-life scenario, the method has to overcome stochasticity and irreversible actions (e.g. breaking a glass) which break that assumption. Also, if an object is moved, you would have to place it back to its original position. However, this could be overcome by learning an accurate forward model of the environment that allows the agent to predict what will happen when it plays an action sequence. Consider the forward model as a proxy for one of the experiences. Recent works have made significant progress in this direction \cite{ha2018recurrent, hafner2020mastering}. Using this forward model, the agent could play one action sequence and then imagine what would have happened if it had played it in a different order, thus performing an SC-experiment. We believe this is an important future work for using sensory commutativity to build perception for artificial agents, drawing links with the processes of visual attention and surprise \cite{itti2006bayesian}.

\quad \textbf{Conclusion.} We studied the sensory commutativity of action sequences for embodied agent scenarios. We introduced SC-experiments and the SCP criterion. We showed that SCP is a good proxy for estimating the effect of each action on the environment, for 2D and 3D realistic embodied scenarios. We illustrated the potential usefulness of such criterion and SC-experiments in general by performing movable and immovable object detection and improving sample-efficiency in an RL problem.



\bibliographystyle{ieeetr}
\bibliography{bibli.bib}

\newpage
\appendix

\section{Alternative methods description}
\label{app:alternatives}

The SCP criterion derived in this paper estimates how much each degree of freedom affects the environment in an embodied agent scenario. In this section we discuss why other approaches cannot reliably estimate the same quantity.

\quad \textbf{Naive approach: changes in sensors.} A straightforward approach to this problem would be to play action sequences of each degree of freedom and quantify how much the sensors change. We consider the squared difference for a transition, i.e. the squared difference for two consecutive observations separated by an action sampled from one dimension of the action space. We report the mean squared difference over 100k transitions, for each degree of freedom. 

It is clear in our experiment results, shown in Fig.~\ref{fig:conj_res_exp}, that the approach fails. For instance, rotating the head of the agent changes dramatically what the agent sees, even though this degree of freedom does not affect the environment. It would have made sense if we had considered the top view (fully-observable scenario), since rotating the head does not changes the top view a lot. However in the embodied scenario, this strategy is not viable. For the same reason, approaches based only on the changes in the embodied sensors are bound to fail.


\quad \textbf{Prediction error approach.} A more involved approach would be to use prediction on the sensory change caused by each degree of freedom, a common approach used to improve exploration in RL \cite{burda2018exploration, pathak2017curiosity}. The DOF that are harder to predict could be the ones affecting the environment the most, and thus being the most important for manipulation and navigation. We tested this alternative in our experiments, by using a feed-forward neural network to predict the next sensor. The neural network takes a concatenation of the sensor and action at time $t$ and predicts the sensor at time $t+1$. We use the same dataset of transitions as in our experiments with the naive baseline (100k transitions for each degree of freedom, 80k for training and 20k for testing). We trained one model for each degree of freedom, using a neural network with two linear hidden layers with the same number of neurons as the input size. We report the excess prediction error on the held-out test set, i.e. the value of the prediction error minus the minimum prediction error among all 8 degrees of freedom. If the method works, higher excess error prediction should indicate a degree of freedom with more effect on the environment.

The results are shown in Fig.~\ref{fig:conj_res_exp}. It turns out that prediction error is not well correlated with how much a degree of freedom is important for navigation and manipulation. For instance, head rotation, which does not affect the environment, is hard to predict: the agent might not know what's outside his field of view. On the contrary, base longitudinal movement affect the environment a lot and is easier to predict than head rotation. 

To conclude, in our experiments we did not find any viable strategy to replace the SCP criterion. SCP is able to easily estimate how important a degree of freedom is for acting and navigating in the environment. The other considered baselines do not manage to organize the action space in the same hierarchical way.

\section{Additional experiments on Flatland}
\label{app:add_exps}

\begin{figure*}[h]
    \centering
    \includegraphics[scale=0.28]{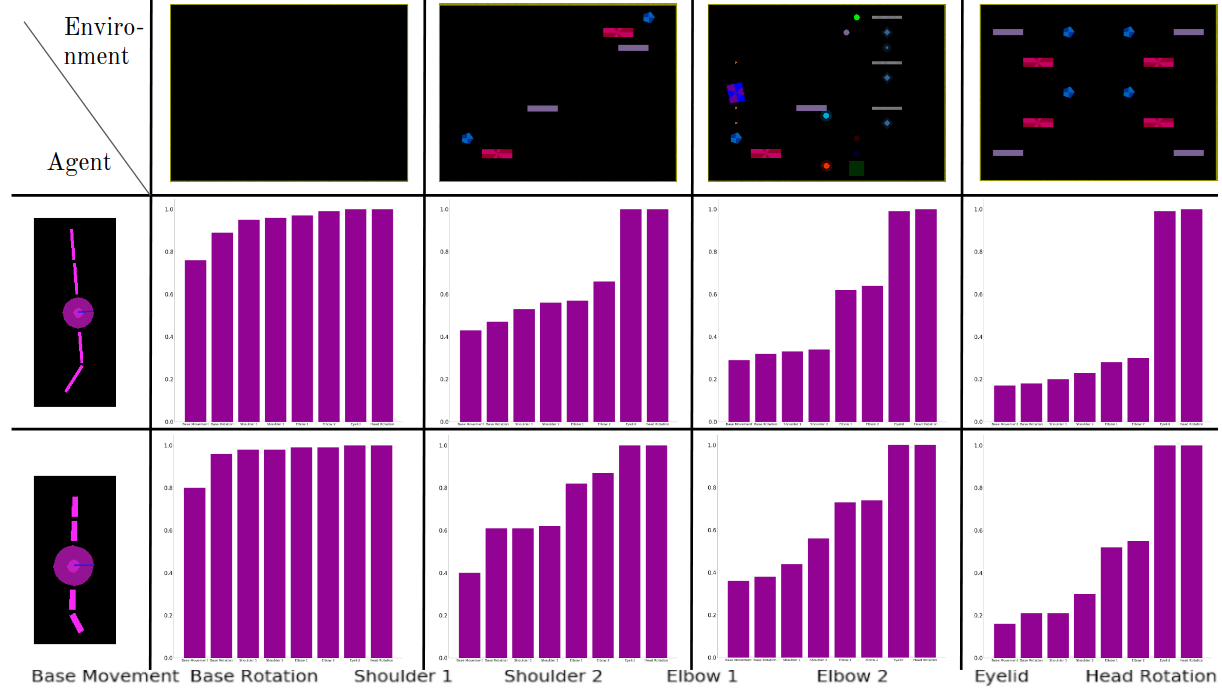}
    \caption{SCP computed for different combinations of agents and environments. Columns: environments. Rows: agents.}
    \label{fig:add_exps}
\end{figure*}

In our additional experiments on Flatland, we verify some of the intuitions we built with the main experiments on Flatland. For that, we compute the SCP as described in Sec.~\ref{sec:scpp} for different combinations of agents and environments. The agents and environments tested are displayed on Fig.~\ref{fig:add_exps}: we use environments with different numbers of objects (from empty to 12 objects), and two agents: one with longer arms than the other.

The results are also displayed on Fig.~\ref{fig:add_exps}. Our intuitions are validated since the more objects are place in the environment, the smaller the value of SCP for DOF that correspond to interacting with these objects. For instance in the empty space almost all DOF have a SCP of 1 since there is nothing to interact with but the walls (that's SCP is not perfectly 1 for base movement annd rotation, shoulder and elbow joints). 

Also, we notice that if the arms are longer, the SCP for shoulder and elbow joints is consistently lower for each environment. Indeed, there is more chance to interact with objects if the arms are longer, thus inducing a lower SCP.

\section{Additional experiments on iGibson}
\label{app:add_exp_igibson}

\begin{figure*}[h]
    \centering
    \includegraphics[scale=0.58]{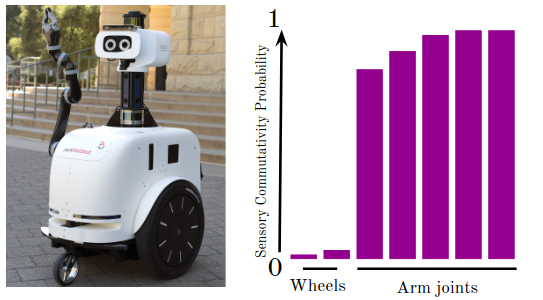}
    \caption{SCP for the JackRabbot (left) in the Rs environment.}
    \label{fig:add_exp_igibson}
\end{figure*}

We follow the same protocol as with the Fetch robot, i.e. we use the Rs environment and the same algorithm to compute the SCP for the 7 degrees of freedom of the JackRabbot: two wheels and a $5$-DOF articulated arm. The results are presented in Fig.~\ref{fig:add_exp_igibson}. We observe the hierarchical organization of the DOF of the agent, the wheels having a low SCP as they allow the robot to move around, and the DOF of the articulated arm having a higher and higher SCP as we move closer to the end of the arm (and thus closer to fine motor skills).

\section{Experimental details for object detection experiments}
\label{app:exp_obj}

We provide further details on the object detection experiments:

\begin{itemize}
    \item The dataset is composed of roughly 10k instances for each possible outcome (identical, completely different, identical up to moved objects).
    \item In order to generate the data for the "completely different" outcome, we apply a 90 degrees rotation to the robot.
    \item For the inference results on movable objects, we experimented with two strategies for the action sequence. Either 20 steps random action sequences or pre-determined action sequences (10 steps where the arm moves left, then 10 steps where the arm moves right).
    \item For the immovable object detection and creation of the map, we use random action sequences of 100 steps.
\end{itemize}

\end{document}